\documentclass{article}
\usepackage[preprint]{style}

\usepackage[utf8]{inputenc} % allow utf-8 input
\usepackage[T1]{fontenc}    % use 8-bit T1 fonts
\usepackage[colorlinks=true]{hyperref}       % hyperlinks
\usepackage{url}            % simple URL typesetting
\usepackage{booktabs}       % professional-quality tables
\usepackage{amsfonts}       % blackboard math symbols
\usepackage{nicefrac}       % compact symbols for 1/2, etc.
\usepackage{microtype}      % microtypography
\usepackage{xcolor}         % colors
\usepackage{algorithm,algorithmic}
\usepackage{wrapfig}

\usepackage{color}
\usepackage{subcaption}
\usepackage{amsmath}
\usepackage{amssymb}
\usepackage{enumitem} 
\usepackage{graphicx}
\usepackage{multirow}
\usepackage{amsmath, mathrsfs}
\usepackage{bbm}
\usepackage{cleveref}
\crefformat{equation}{Eq.~#2#1#3}
\Crefformat{equation}{Equation~#2#1#3}
\usepackage{colortbl}
\definecolor{myblue}{RGB}{218,232,252}

\title{SDMPrune: Self-Distillation MLP Pruning for Efficient Large Language Models}

\author{%
  Hourun Zhu\\
  Department of Computer Science\\
  Central South University\\
  \texttt{zhuhr@csu.edu.cn} \\
  \And
  Chengchao Shen\thanks{Corresponding Author}\\
  Department of Computer Science\\
  Central South University\\
  \texttt{scc.cs@csu.edu.cn} \\
}

\begin{document}

\maketitle

\begin{abstract}
    In spite of strong performance achieved by LLMs, the costs of their deployment are unaffordable. 
    For the compression of LLMs, gradient-based pruning methods present promising effectiveness.
    However, in these methods, the gradient computation with one-hot labels ignore the potential predictions on other words, thus missing key information for generative capability of the original model. 
    To address this issue, we introduce a self-distillation loss during the pruning phase (rather than post-training) to fully exploit the predictions of the original model, thereby obtaining more accurate gradient information for pruning.
    Moreover, we find that, compared to attention modules, the predictions of LLM are less sensitive to multilayer perceptron (MLP) modules, which take up more than $5 \times$ parameters (LLaMA3.2-1.2B).
    To this end, we focus on the pruning of MLP modules, to significantly compress LLM without obvious performance degradation.
    Experimental results on extensive zero-shot benchmarks demonstrate that our method significantly outperforms existing pruning methods.
    Furthermore, our method achieves very competitive performance among 1B-scale open source LLMs.
    The source code and trained weights are available at \url{https://github.com/visresearch/SDMPrune}.
\end{abstract}

\section{Introduction}
\label{sec:introduction}

Large Language Models~\cite{openai2024gpt4technicalreport,touvron2023llamaopenefficientfoundation,bai2023qwentechnicalreport,deepseekai2024deepseekv3technicalreport} have demonstrated remarkable performance across various natural language processing tasks, with empirical evidence showing a positive correlation between model capabilities and increasing parameter count.
However, excessive computing and memory costs greatly limit their economic deployment on widespread applications.
Extensive research has shown that model structural pruning, a key approach in model compression, is particularly effective in addressing this challenge.

\begin{figure}[t]
  \centering
  \begin{subfigure}{0.48\textwidth}  % 调整为48%的文本宽度
      \centering
      \includegraphics[width=\linewidth]{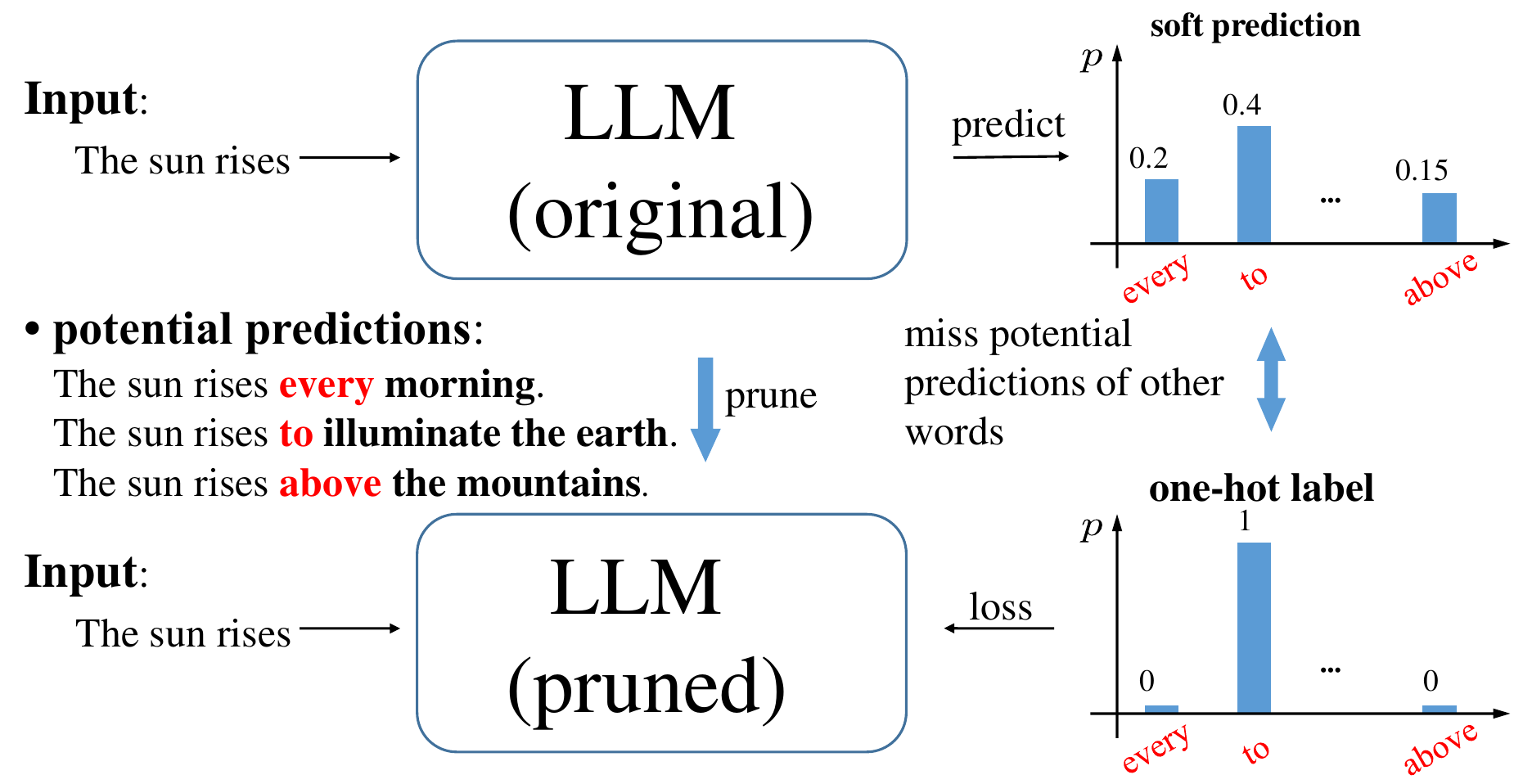}
      \caption{Self-distillation loss vs One-hot label loss}
      \label{fig:self-distillation}
  \end{subfigure}
  \hfill  % 添加水平填充使两图分开
  \begin{subfigure}{0.48\textwidth}  % 调整为48%的文本宽度
      \centering
      \includegraphics[width=\linewidth]{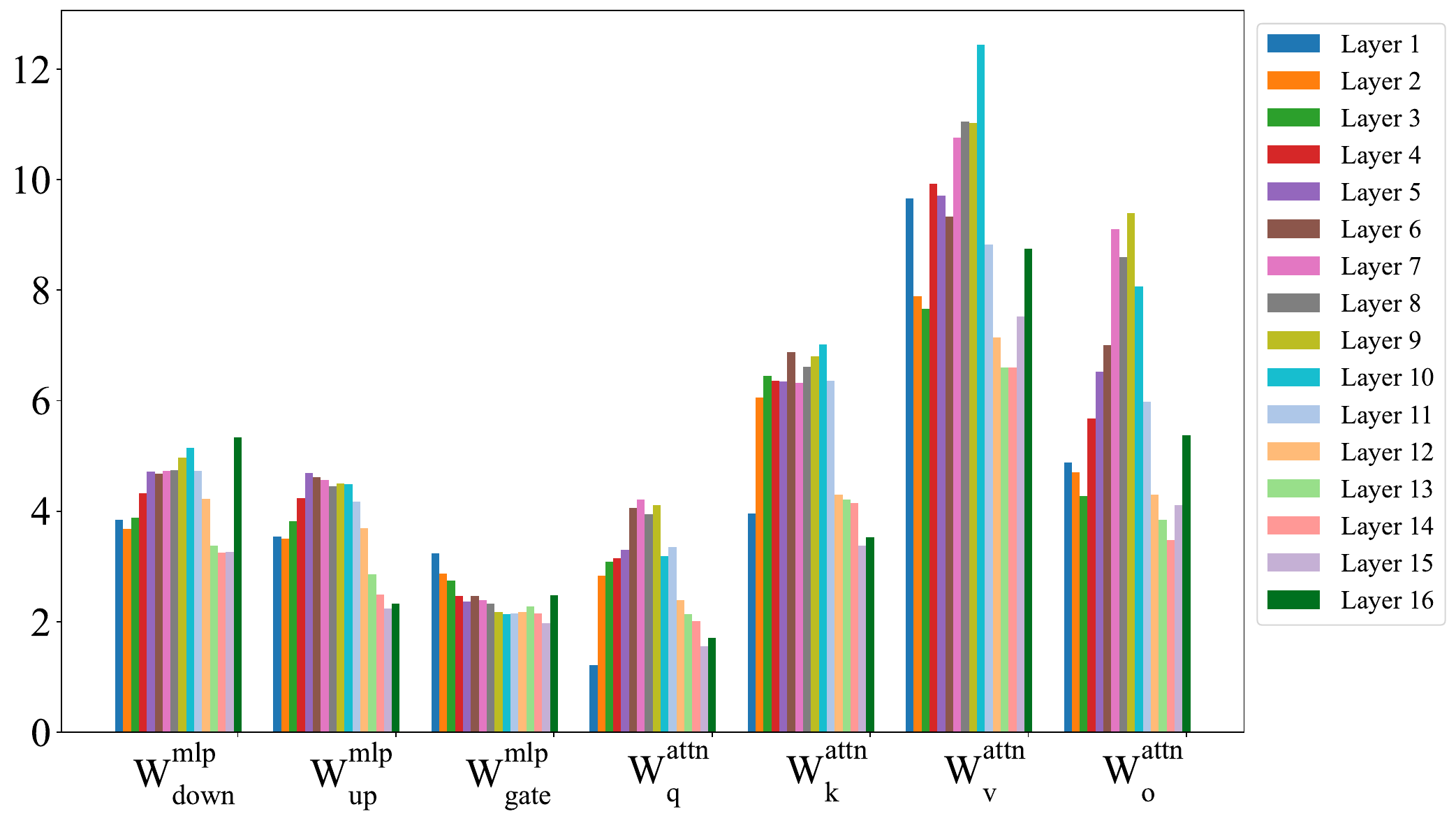}
      \caption{Average importance scores in linear params of MLP and attention layers.}
      \label{fig:motivation_}
  \end{subfigure}
  \caption{
  (a) Using the complete prediction probability distribution of the original model as the pruning objective leads to models with superior performance. 
  (b) Following the Taylor pruning criterion\cite{molchanov2016pruning,molchanov2019importance}, the importance scores in this figure were computed using the LLaMA3.2-1.2B model on the C4 dataset.
  }
  \label{fig:motivation}
  \vspace{-1em}
\end{figure}

In the field of model structural pruning research, gradient-based structural pruning methods \cite{das2024sizegradientsshapepruning} have demonstrated superior performance applied to Transformer architectures across various domains. 
A notable example is the Taylor pruning method \cite{molchanov2016pruning}, which has shown particular effectiveness.
In Taylor pruning, importance scores are computed as the element-wise product of parameters and their corresponding gradients.
However, as shown in Figure \ref{fig:self-distillation}, existing pruning methods predominantly obtain the importance scores by one-hot label encoding loss, which ignores the potential predictions of other words, thus missing vital 
information for the preservation of the original generative capability.
Furthermore, structural pruned model typically requires one or more rounds of self-distillation post-training to reach usable performance levels which incurs significant computational overhead.
To address this limitation, we integrate distillation post-training constraints into pruning process by employing self-distillation loss function and evaluating neuron importance scores using a Taylor expansion-based metric. 
% our pruning methodology focuses predominantly on preserving the complete predictive probability distribution of the original model. 
% We employ a self-distillation objective function and evaluate neuron importance scores using a Taylor expansion-based metric. 
This approach effectively mitigates the common pruning drawback of neglecting potential predictive vocabulary and maintains the original model's semantic analysis capabilities.
In this manner, the derived gradient-based importance scores preferentially prune neurons exhibiting lower sensitivity to the original model's predictions which can minimize performance degradation.

We further analyze which part of LLM to prune.
To this end, we evaluate the Taylor-based importance scores of all attention and MLP modules of LLaMA3.2-1.2B \cite{grattafiori2024llama3herdmodels} as Figure \ref{fig:motivation_}.
We find that the average importance scores of the parameters in MLP modules are substantially lower than the ones in attention modules.
It indicates that MLP modules contribute less to the predictions of model than the attention modules.
However, MLP modules (such as LLaMA3.2-1.2B) take up more than $5\times$ parameters of the attention modules.
Consequently, pruning MLP modules offer a more efficient solution to LLM compression, effectively reducing the model size while minimizing performance loss.
Specifically, due to the large hidden dimension size, we focus on the pruning of hidden layer neurons of MLPs.
Meanwhile, the pruning of hidden layers doesn't change the output size of MLPs, thereby having less effect on the structure of the pruned LLM and more enhanced compatibility with the original model for better performance recoverability.

The contributions of our work can be summarized as follows.
\begin{itemize}[leftmargin=0pt, itemindent=*, noitemsep]
  \item We propose a self-distillation pruning method to obtain more accurate gradient-based importance scores by fully utilizing the predictions of the original model.
  \item We find that the predictions of LLMs are less sensitive to MLP modules than attention ones.
  The experimental results also support that only pruning MLP modules can significantly reduce the parameters of LLMs, but with less performance degradation compared to the pruning of attention modules.
  \item Our method significantly outperforms the existing LLM pruning methods.
  Moreover, our method achieves a very competitive performance among 1B-scale open source LLMs using only a small amount of data for finetuning.
\end{itemize}
\section{Related Work}
\label{sec:related_work}

\textbf{Self-distillation methods}.
Previous research~\cite{zhang2019your,pham2022revisitingselfdistillation,shen2021progressive,shen2019amalgamating} has demonstrated that self-distillation training serves as an effective method for enhancing model performance.
In the domain of LLM, self-distillation has also been extensively adopted in their training and optimization processes.
Open-source LLM series like LLaMA typically employ a methodology where full-parameter foundation models are first trained, then optimized through structural pruning and self-distillation to produce efficient and compact models.
Minitron~\cite{muralidharan2024compact} introduces an iterative pruning-self-distillation framework for model compression. 
However, each pruning iteration necessitates a complete self-distillation retraining process, resulting in multiplicative growth of total computational costs.
To address this limitation, we integrate the distillation process within the pruning process and we can obtain the high-performance model compression through single-pass pruning-training. 

\textbf{Local-Invariant compression methods}.
Prior research \cite{NIPS1989_6c9882bb,hassibi1993optimal} proposed a pruning optimization framework that analyzes parameter importance while preserving layer-wise output consistency, though requiring second-order derivative information.
Subsequently, OBC \cite{frantar2023optimalbraincompressionframework} significantly reduces computational overhead by performing calculations and accumulating compression errors at the row granularity. 
GPTQ \cite{frantar2023gptqaccurateposttrainingquantization} accelerates this process through arbitrary-order quantization, lazy batch updates and Cholesky decomposition-based reconstruction.
SparseGPT \cite{frantar2023sparsegptmassivelanguagemodels} and SlimGPT ~cite{ling2024slimgpt} have successfully applied this approach to sparse pruning and structural pruning in LLM.
These methods suffer from a severe computational bottleneck due to the Hessian matrix computation which makes these methods difficult to apply in practice.

\textbf{Global Gradient compression methods}.
NVIDIA's works \cite{molchanov2016pruning,molchanov2019importance} aims to minimize the loss change of the pruned model and utilizes first-order Taylor expansion for model pruning.
LLM-Pruner \cite{ma2023llm} treats the model as a group and employs first-order importance to evaluate parameter significance. 
LoRAPrune \cite{zhang2024loraprunestructuredpruningmeets} evaluates the importance of weights based on the gradients of LoRA parameters instead of the model's parameters, also demonstrating commendable performance.
These methods rely only on label information while neglecting more critical knowledge from the original model, leading to erroneous pruning substructures.

\textbf{Other pruning methods}.
Magnitude \cite{han2015learningweightsconnectionsefficient} pruning uses the model weights as importance scores for pruning.
Building on this, Wanda \cite{sun2024simpleeffectivepruningapproach} employs the product of weights and activation values as importance scores.
Sheared-LLaMA \cite{xia2024shearedllamaacceleratinglanguage} employs enhanced $\ell_0$ regularization on masks to simultaneously prune both the width and depth of the model.
Compresso \cite{guo2023compressostructuredpruningcollaborative} integrates $\ell_0$ regularization with LoRA training and perform collaborative prompt-guided pruning to compress the width of the model.
SliceGPT \cite{ashkboos2024slicegptcompresslargelanguage} achieves pruning by applying orthogonal matrix transformations to ensure computational invariance.
OWL \cite{yin2024outlierweighedlayerwisesparsity} determines the pruning ratio for each layer based on its layer-wise outlier distribution, where the pruning ratio is inversely proportional to the proportion of outlier features in that layer.
LLaVA-STF~\cite{tang2025compact} shortens the length of input sequences by reducing the spatial redundancy among vision tokens, thus alleviating the quadratic complexity computational challenge suffered by large multimodal models.
% These pruning methods, while contributing valuable insights to pruning research, exhibit limited generalizability and inconsistent performance across diverse model architectures.
\section{Preliminary}
\label{sec:preliminary}

Oracle pruning method~\cite{NIPS1988_07e1cd7d} minimizes performance degradation in pruned models by selectively removing weights whose removal incurs the least increased error which can be expressed as:
\vspace{-0.5em}

\begin{equation}
  \min_{W^{\prime}} \left| \mathcal{L}(D|W^{\prime}) - \mathcal{L}(D|W) \right|
\end{equation}

where \( W \) represents the original model parameters, \( W' \) denotes the pruned model parameters and \( D \) denotes the calibration dataset.

As shown in Figure ~\ref{fig:motivation_}, due to the substantially lower importance scores and dominant parameter count of MLP modules, our method primarily targets MLP parameters for pruning.
% In transformer architectures, MLP modules are conventionally implemented as a sequence of linear layers. 
The LLaMA model family, for instance, employs a three-layer MLP structure consisting of upsample linear, gate linear and down-sample linear.
We denote the parameters of these linear layers as $W_u$, $W_g$, $W_d$, respectively. 
Considering the direct row-column correspondence between parameters in the three linear layers, $W_u^{(i)}$, $W_g^{(i)}$ and $W_d^{(i)}$ represent the parameters associated with the \( i \)-th hidden layer neuron.
For structural pruning, the importance score of each hidden layer neuron in MLP modules can be represented as \cref{eq:oracle_mlp}.
\vspace{-0.5em}

\begin{align}
  I_i &= \left|\mathcal{L}\left(D|W_d^{(i)}=0,W_g^{(i)}=0,W_u^{(i)}=0\right) - \mathcal{L}\left(D|W_d^{(i)},W_g^{(i)},W_u^{(i)}\right)\right|
  \label{eq:oracle_mlp}
\end{align}

% To compute \cref{eq:oracle_mlp} , most existing approaches employ Taylor expansion approximations.
To achieve a balance between model accuracy and computational cost, we employ an approximate computation based on first-order Taylor expansion.
Inspired by Taylor pruning ~\cite{molchanov2016pruning,molchanov2019importance} , we formulate this problem as a multivariate optimization function and employ first-order Taylor expansion to approximate the importance evaluation for each hidden neuron in MLP modules.
The formula can be simplified as \cref{eq:1taylor}:
% \begin{figure*}[t]
%   \begin{equation}
%   \begin{aligned}
%     &\mathcal{L}(D|W_d^{(i)}=0,W_g^{(i)}=0,W_u^{(i)}=0) \\
%     & = \mathcal{L}(D|W_d^{(i)},W_g^{(i)},W_u^{(i)})- \frac{\partial \mathcal{L}}{\partial W_d^{(i)}} \cdot W_d^{(i)} - \frac{\partial \mathcal{L}}{\partial W_g^{(i)}} \cdot W_g^{(i)} - \frac{\partial \mathcal{L}}{\partial W_u^{(i)}} \cdot W_u^{(i)} \\
%     & + R_2(W_d^{(i)}=0,W_g^{(i)}=0,W_u^{(i)}=0)
%   \end{aligned}
%   \label{eq:2taylor}
%   \end{equation}
% \end{figure*}

\begin{equation}
  \begin{aligned}
    &\mathcal{L}(D|W_d^{(i)}=0, W_g^{(i)}=0, W_u^{(i)}=0) \\
    &\approx \mathcal{L}(D|W_d^{(i)}, W_g^{(i)}, W_u^{(i)}) - \frac{\partial \mathcal{L}}{\partial W_d^{(i)}} \cdot W_d^{(i)} - \frac{\partial \mathcal{L}}{\partial W_g^{(i)}} \cdot W_g^{(i)} - \frac{\partial \mathcal{L}}{\partial W_u^{(i)}} \cdot W_u^{(i)}
  \end{aligned}
  \label{eq:1taylor}
\end{equation}

% \cref{eq:1taylor} can be easily computed because the gradients have already been obtained during the backpropagation process. 
% Through such calculations, we derive the importance scores for each hidden layer neuron, which quantify the contribution of individual neurons to the overall model performance. 
In this way, we derive the importance scores for each hidden layer neuron, which quantify the contribution of individual neurons to the overall model performance. 
The final formula can be represented as \cref{eq:importance}
% \begin{align}
%   I_i &\approx \left| \frac{\partial \mathcal{L}}{\partial W_d^{(i)}} \cdot W_d^{(i)} + \frac{\partial \mathcal{L}}{\partial W_g^{(i)}} \cdot W_g^{(i)}  \right.  \nonumber  \\
%       &\left. \quad \quad \quad \quad + \frac{\partial \mathcal{L}}{\partial W_u^{(i)}} \cdot W_u^{(i)} \right|
%   \label{eq:importance}
% \end{align}
\vspace{-0.5em}

\begin{small}
  \begin{equation}
    I_i \! \approx \! \left| \frac{\partial \mathcal{L}}{\partial W_d^{(i)}} \! \cdot \! W_d^{(i)} \! + \! \frac{\partial \mathcal{L}}{\partial W_g^{(i)}} \cdot W_g^{(i)} \! + \! \frac{\partial \mathcal{L}}{\partial W_u^{(i)}} \cdot W_u^{(i)} \right|
    \label{eq:importance}
  \end{equation}
\end{small}
\section{Method}
\label{sec:method}

% This section details our novel self-distillation framework for structured pruning of MLP layers in large language models.
% The methodology is delineated into three principal components:
% \textbf{Section3.1} elucidates the method for estimating the importance of hidden neurons.
% \textbf{Section3.2} analyzes the self-distillation structured pruning.
% \textbf{Section3.3} delineates the structured pruning procedure.
\subsection{Overview}

\begin{figure*}[ht]
    \centering
    \includegraphics[width=\linewidth]{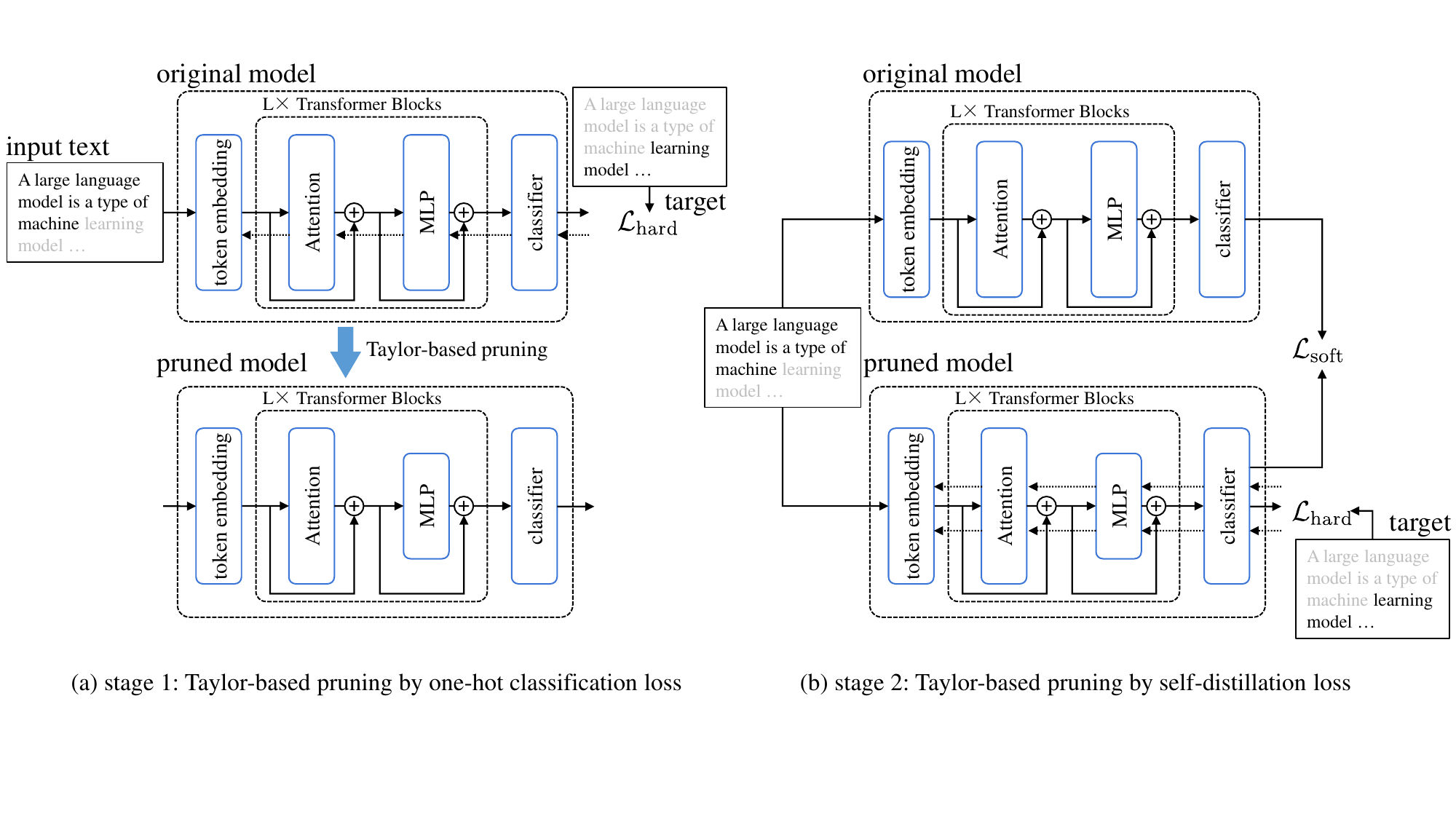}
    \vspace{-1.5cm}
    \caption{Overview of the SDMPrune.}
    \label{fig:method}
\end{figure*}

To efficiently compress the model, we introduce self-distillation loss for accurate evaluation of importance weights in MLP hidden-layer neurons.
As depicted in Figure ~\ref{fig:method}, we adopt a dual stage model compression strategy.
In the first stage, since no pruned model has been generated yet, it is impossible to directly construct a distillation loss with the original model. 
Therefore, we establish preliminary model parameter differences based on one-hot classification loss in a low pruning retio.
In the second stage, we perform thorough model pruning using the self-distillation loss where the original model is regarded as the teacher.

\subsection{Self-Distillation Loss}

In existing works, gradient information was obtained by computing the one-hot label encoding loss, which is used to enhance classification accuracy. 
However, these methods overlook the potential prediction of other words except the label and compromise the model's ability to generate diverse outputs.
To address this issue, we introduce a self-distillation loss, which fully exploits the potential predictions of the original model, enabling effective pruning while maintaining the model's generative capabilities.

During the initial pruning phase, the pruned model is same as the original model which renders the computation of self-distillation loss mathematically infeasible.
Therefore, we divide the pruning process as two distinct stages.

\textbf{Label-based cold-start stage}, as the first stage, aligns with previous methodologies. 
In this stage, we optimize the calibration dataset's generation accuracy through cross-entropy loss.
We implement an initial pruning phase to establish parameter-space divergence, thereby enabling computation of self-distillation loss.

\textbf{Self-distillation loss based calibration stage}, as the second stage, involves using the original model as the teacher and the model obtained from the first stage as the student. 
By applying distillation loss on the calibration data, we prune the model with the goal of preserving the original model's full linguistic capabilities. 
This approach ensures that the pruned model retains both the essential predictive ability and the nuanced language understanding of the original, unpruned model.
We employ \cref{eq:loss} as the self-distillation loss function.
\begin{equation}
  \mathcal{L}_{\text{dis}} = (1 - \alpha) \cdot \mathcal{L}_{\text{hard}} + \alpha \cdot \mathcal{L}_{\text{soft}}
  \label{eq:loss}
\end{equation}

The distillation loss consists of two components: the \textbf{hard loss} and the \textbf{soft loss}, as defined as \cref{eq:hardloss} and \cref{eq:softloss}.
The soft loss measures the KL divergence between the teacher and student models' output distributions, ensuring the pruned model retains the original's comprehensive language generation capabilities. 
The hard loss calculates crossentropy loss on calibration dataset, guaranteeing accurate performance on specific tasks.
Dynamic interpolation between these objectives yields pruned models with both robust generalization and sharp task accuracy.

\begin{equation}
  \mathcal{L}_{\text{hard}} = \sum_{i=1}^V \mathbbm{1}_{i=y} \cdot \log(q_i)  
  \label{eq:hardloss}
\end{equation}  
\begin{equation}
  \mathcal{L}_{\text{soft}} = \sum_{i=1}^V (p_i/T) \cdot \log(\frac{q_i}{p_i/T})  
  \label{eq:softloss}
\end{equation}  

where \( p_i \) is the predicted probability distribution output by the teacher model for class \( i \), \( q_i \) is the predicted probability distribution output by the student model for class \( i \), \( V \) is the size of the model's vocabulary.

\subsection{Structural Pruning}

In the context of structural pruning, since the neurons being pruned are associated with multiple weight matrices, it is essential to consider the interconnections between multiple linear layers. 
Unlike sparsification pruning, which targets individual neurons in isolation, structural pruning involves setting entire rows or columns of parameters to zero in the MLP.

\begin{figure}[htbp]
  \centering
  \includegraphics[width=0.4\textwidth]{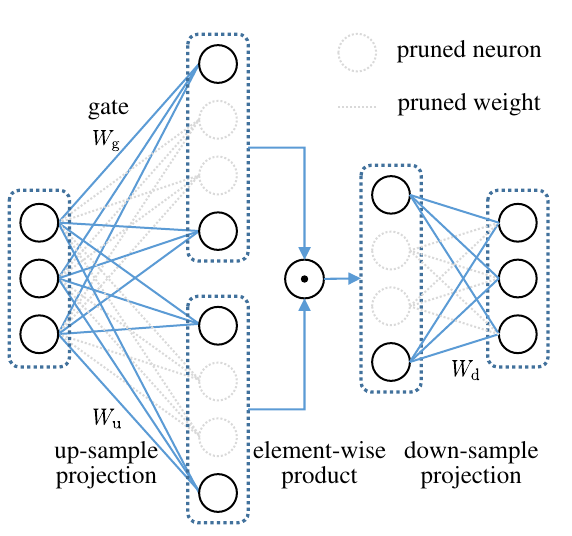}
  \caption{Taking LLaMA MLP as an example, this describes the structural pruning of MLP.}
  \label{fig:structural_prune}
\end{figure}

After calculating importance scores for each neuron in the hidden layer, the neurons are ranked and the retained subset is selected according to \cref{eq:indices}.
As depicted in Figure \ref{fig:structural_prune}, the parameters associated with pruned neurons are zeroed out and removed. 
This operation effectively reduces the intermediate dimensionality of the MLP, thereby optimizing its structure and efficiency.

\begin{equation}
  S = \{ i \mid I_i \in \text{top}_k(\{I_1, I_2, \dots, I_N\}) \}
\label{eq:indices}
\end{equation}
The overall algorithm is summarized as Algorithm~\ref{alg:alg}.

\renewcommand{\algorithmicrequire}{\textbf{Input:}}
\renewcommand{\algorithmicensure}{\textbf{Output:}}
\begin{algorithm}[ht]
  \caption{Self-Distillation Structural Pruning For MLP}
  \label{alg:alg}
  \begin{algorithmic}[1]
    \REQUIRE{
          Calibration dataset $D$; Down\_proj linear weight: \( W_d \); Gate\_proj linear weight: \( W_g \); Up\_proj linear weight: \( W_u \); Cold-Start Number of Retained Neurons \( k^\prime \); Target Number of Retained Neurons \( k \); 
          Hard loss function $\mathcal{L}_{hard}$; Self-distillation loss function $\mathcal{L}_{dis}$}.
    \ENSURE{
          Pruned weights $W_\mathrm{d}^\mathrm{\prime}$, $W_\mathrm{g}^\mathrm{\prime}$ and $W_\mathrm{u}^\mathrm{\prime}$.}
      
      \FOR{each data \( x \) in \( D \)}
        \STATE Calculate $\mathcal{L}_{hard}$ and backward; 
        \STATE Calculate $I_i$ via \cref{eq:importance};
      \ENDFOR
      \STATE Select top $k^\prime$ neurons via \cref{eq:indices};
      \STATE Obtain $W_\mathrm{d}^\mathrm{stage1}$, $W_\mathrm{g}^\mathrm{stage1}$ and $W_\mathrm{u}^\mathrm{stage1}$

      \FOR{each data \( x \) in \( D \)}
        \STATE Calculate $\mathcal{L}_{dis}$ and backward;
        \STATE Calculate $I_i$ via \cref{eq:importance};
      \ENDFOR
      \STATE Select top $k$ neurons via \cref{eq:indices};
      \STATE Obtain $W_\mathrm{d}^\mathrm{\prime}$, $W_\mathrm{g}^\mathrm{\prime}$ and $W_\mathrm{u}^\mathrm{\prime}$
  \end{algorithmic}
\end{algorithm}

\section{Experiment}
\label{sec:experiment}
\subsection{Experimental Setup}
\textbf{Model and Metrics}.
Our structural pruning method is applied to the LLaMA3.2 model and LLaMA2 model. 
Following the setup of previous work~\cite{ma2023llm, zhang2024loraprunestructuredpruningmeets}, we evaluate perplexity on WikiText \cite{merity2016pointer} and the zero-shot performance of the model is assessed using lm\_eval \cite{eval-harness}, covering datasets such as 
ARC-easy \cite{allenai:arc}, ARC-challenge \cite{allenai:arc}, BOOLQ \cite{clark2019boolqexploringsurprisingdifficulty}, Crows-Pairs \cite{nangia2020crowspairschallengedatasetmeasuring},
OpenBookQA \cite{mihaylov2018suitarmorconductelectricity},PIQA \cite{bisk2019piqareasoningphysicalcommonsense},
Race \cite{lai2017racelargescalereadingcomprehension}, SocialQA \cite{sap2019socialiqa}, TruthfulQA \cite{lin2021truthfulqa},
Winogrande \cite{sakaguchi2019winograndeadversarialwinogradschema}.

\textbf{Calibration Dataset}.
We employ the C4 dataset \cite{raffel2020exploring} as the calibration dataset, utilizing 1024 randomly sampled sequences with a fixed length of 1024.
The sampling approach is identical to that of wanda \cite{sun2024simpleeffectivepruningapproach}, where for each sample exceeding the sequence length of 1024, a random segment of 1024 tokens is extracted. 
In Section ~\ref{sec:comparewithmethods}, Section ~\ref{sec:comparewithmodels} and Section ~\ref{sec:ablation},
to align the experimental configurations, We finetuned the model on the Lamini-instruction dataset \cite{wu2023lamini}, which was generated from instructions and responses based on several existing prompt resources and multiple datasets using GPT-3.5-turbo.

\textbf{Implementation Details}.
All experiments are conducted using the AdamW optimizer with BF16 precision, and the learning rate is following a cosine scheduler.
Section ~\ref{sec:comparewithmethods} and ~\ref{sec:ablation} employ LoRA for model finetuning and Section ~\ref{sec:comparewithmodels} utilizes full-parameter finetuning.
More implementation details can be found in the Appendix.
% \begin{figure}[htbp]
%   \includegraphics[width=0.48\textwidth]{pack.png}
%   \caption{Visualization of the attention mask for packed samples.The attention mask maintains the relationships within each sample while ensuring that the samples are mutually invisible to each other.}
%   \label{fig:pack}
% \end{figure}
\begin{table*}[ht]
    \centering
    \caption{Zero-shot performance and Perplexity of the compressed LLaMA3.2-1.2B model, which are finetuned on the LaMini dataset using LoRA for 2 epochs. 
    We evaluate perplexity on WikiText2 with 512-token segments. 
    The average accuracy is calculated among ten classification datasets. 
    \textbf{Bold} denotes the best performance at the same compression rate.
    % The results highlighted in blue background represent the outcomes of our proposed methodology.
    The pruning ratio refers to the proportion of pruned parameters to the total parameters, \textbf{including the embedding layer}. 
    % The model parameter count obtained at the three different ratios are 0.989B, 0.865B, and 0.742B, respectively.
    The parameters of pruned models with pruning ratios 20\%, 30\% and 40\% are 0.989B, 0.865B and 0.742B, respectively.
    }
    \resizebox{\textwidth}{!}
    {
    \begin{tabular}{l|l|c|ccccccccccc}
        \hline
        \textbf{Ratio}
        & \textbf{Method} & \textbf{PPL$\downarrow$}
        & \textbf{ARCc} & \textbf{ARCe} & \textbf{BOOLQ} &\textbf{Crows} & \textbf{OBQA}
        & \textbf{PIQA} & \textbf{Race} & \textbf{SiQA} & \textbf{TfQA}  & \textbf{Wino}  
        & \textbf{Average$\uparrow$}
        \\
        \hline\hline
        0\% & LLaMA3.2-1.2B&12.98&37.12&60.69&64.04&62.55&37.6&74.16&37.61&42.89&37.70&60.38&51.47 \\
        \hline\hline
        \multirow{6}{*}{20\%} 
        & Magnitude \cite{han2015learningweightsconnectionsefficient} &36.01 &25.35 
        &47.33 &58.07 &56.93 &30.4 &66.33 &32.03 &41.21&40.12&53.46&45.12 \\
        & Wanda \cite{sun2024simpleeffectivepruningapproach}
        &33.27 &28.33 &50.23 &62.95 &54.98 &28.9 &65.89 &33.84 &41.00 &\textbf{42.32} &54.23 &46.27 \\
        & LLMPruner \cite{ma2023llm}
        &28.92 &29.61 &51.00 &62.07 &57.54 &\textbf{33.6} &67.19 &33.68 &40.89 &41.68 &56.59 &47.39 \\
        & Compresso \cite{guo2023compressostructuredpruningcollaborative}
        &- &27.05 &48.53 &59.14 &\textbf{58.56} &27.4 &66.97 &33.21 &40.23 &43.74 &55.64 &46.05 \\
        & LoRAPrune \cite{zhang2024loraprunestructuredpruningmeets}
        &\textbf{26.68} &29.60 &50.99 &66.18 &54.38 &29.6 &66.36 &34.05 &42.02 &41.66 &\textbf{56.91} &47.17 \\
        & \cellcolor{myblue}SDMPrune(ours)
        &\cellcolor{myblue}26.96 &\cellcolor{myblue}\textbf{31.14} &\cellcolor{myblue}\textbf{55.22} &\cellcolor{myblue}\textbf{67.58} &\cellcolor{myblue}57.84 &\cellcolor{myblue}32.4 &\cellcolor{myblue}\textbf{70.29} &\cellcolor{myblue}\textbf{35.41} &\cellcolor{myblue}\textbf{42.73} &\cellcolor{myblue}40.20 &\cellcolor{myblue}56.59 &\cellcolor{myblue}\textbf{48.94} \\
        \hline\hline
        \multirow{6}{*}{30\%} 
        & Magnitude \cite{han2015learningweightsconnectionsefficient}
        &47.13 &24.99 &44.83 
        &58.65 &55.58 &26.8 &62.79 &31.25 &38.24 &39.27 &50.32 &43.27  \\
        & Wanda \cite{sun2024simpleeffectivepruningapproach}
        &45.73 &25.35 &45.62&59.02 &52.32 &27.3 &63.19 &31.73 &37.73 &43.12 &54.65 &44.01 \\
        & LLMPruner \cite{ma2023llm}
        &42.46&27.30 &46.34 &60.18 &54.32 &27.4 &63.98 &32.87 &38.43 &\textbf{42.48} &54.64 &44.79 \\
        & Compresso \cite{guo2023compressostructuredpruningcollaborative}
        &- &25.73 &45.33 &61.75 &55.82 &26.5 &62.29 &31.83 &37.13 &40.73 &53.45 &44.06 \\
        & LoRAPrune \cite{zhang2024loraprunestructuredpruningmeets}
        &40.11 &28.07 &46.38 &59.61 &56.29 &27.6 &64.92 &32.37 &37.32 &41.59 &54.38 &44.86 \\
        & \cellcolor{myblue}SDMPrune(ours)
        &\cellcolor{myblue}\textbf{39.70} &\cellcolor{myblue}\textbf{28.50} &\cellcolor{myblue}\textbf{47.47} &\cellcolor{myblue}\textbf{64.68} &\cellcolor{myblue}\textbf{56.89} &\cellcolor{myblue}\textbf{29.0} &\cellcolor{myblue}\textbf{66.32} &\cellcolor{myblue}\textbf{33.21} &\cellcolor{myblue}\textbf{40.84} &\cellcolor{myblue}42.35 &\cellcolor{myblue}\textbf{54.70} &\cellcolor{myblue}\textbf{46.40} \\
        \hline\hline
        \multirow{6}{*}{40\%} 
        & Magnitude \cite{han2015learningweightsconnectionsefficient}
        &100.66 &24.18 &40.37 
        &56.12 &51.42 &25.0 &61.07 &31.24 &37.25 &43.01 &49.81 &41.95  \\
        & Wanda \cite{sun2024simpleeffectivepruningapproach}
        &90.03 &23.54 &39.12&55.13 &\textbf{55.68} &24.2 &60.12 &30.79 &37.92 &44.13 &50.14 &42.08 \\
        & LLMPruner \cite{ma2023llm}
        &76.7 &25.09 &39.94 &58.47 &52.06 &\textbf{28.0} &60.45 &30.33 &38.69 &\textbf{44.90} &51.14 &42.91 \\
        & Compresso \cite{guo2023compressostructuredpruningcollaborative}
        &- &24.35 &40.13 &59.01 &54.09 &26.1 &61.92 &30.61 &37.64 &42.53 &51.09 &42.75 \\
        & LoRAPrune \cite{zhang2024loraprunestructuredpruningmeets}
        &\textbf{68.63} &24.57 &\textbf{44.36} &60.73 &54.32 &24.4 &60.50 &28.52 &37.87 &41.38 &52.33 &42.90 \\
        & \cellcolor{myblue}SDMPrune (ours)
        &\cellcolor{myblue}70.12 &\cellcolor{myblue}\textbf{26.02} &\cellcolor{myblue}42.63 &\cellcolor{myblue}\textbf{65.38} &\cellcolor{myblue}52.59 &\cellcolor{myblue}25.6 &\cellcolor{myblue}\textbf{63.44} &\cellcolor{myblue}\textbf{32.25} &\cellcolor{myblue}\textbf{38.74} &\cellcolor{myblue}43.30 &\cellcolor{myblue}\textbf{52.17} &\cellcolor{myblue}\textbf{44.21} \\
        \hline\hline
      \end{tabular}
    }
    \label{tab:methods}
    % \vspace{-1em}
\end{table*}

% \begin{table}[htbp]
%   \centering
%   \caption{Zero-shot performance and Perplexity of the compressed LLaMA3.2-3B and LLaMA2-7B models.
%   Experiment setting is same as Table \ref*{tab:methods}.Detailed results are available in the Appendix.}
%   \resizebox{\textwidth}{!}
%   {
%   \begin{tabular}{lccccccc}
%   \hline
%   Prune\%& Method & \#Params & PPL $\downarrow$ & Zero-shot A Avg. $\uparrow$ & \#Params & PPL & Zero-shot A Avg. $\uparrow$ \\
%   \hline
%   - & - & 3.2B & 10.26  & 54.30  & 6.7B & 7.18  & 58.33  \\
%   \hline
%   \multirow{3}{*}{20\%}
%   &LLMPruner& 2.56B & 22.07 & 52.24& 5.39B & 11.59 & 56.14\\
%   &LoRAPrune & 2.57B & 19.03 & 50.30 & 5.39B & 11.02 & 56.69 \\
%   &\cellcolor{myblue}SDMPrune & \cellcolor{myblue}2.56B & 
%   \cellcolor{myblue}\textbf{18.8}  & \cellcolor{myblue}\textbf{53.37} & 
%   \cellcolor{myblue}5.38B & \cellcolor{myblue}\textbf{10.48} & \cellcolor{myblue}\textbf{58.21} \\
%   \hline
%   \multirow{3}{*}{30\%}
%   &LLMPruner & 2.25B & 30.12 & 48.68 & 4.71B & 14.76 & 51.69 \\
%   &LoRAPrune & 2.25B & 27.31 & 48.36 & 4.72B & \textbf{12.81} & 54.62 \\
%   &\cellcolor{myblue}SDMPrune & \cellcolor{myblue}2.25B & \cellcolor{myblue}\textbf{27.24} & 
%   \cellcolor{myblue}\textbf{51.03} & \cellcolor{myblue}4.71B & \cellcolor{myblue}12.87 & \cellcolor{myblue}\textbf{55.57} \\
%   \hline
%   \end{tabular}
%   }
%   \label{tab:methods2}
% \end{table}

\begin{table}[htbp]
  \centering
  \caption{Zero-shot performance and Perplexity of the compressed LLaMA3.2-3.2B (left) and LLaMA2-7B (right) models. Experiment setting is consistent with Table \ref*{tab:methods}. Detailed results are available in the Appendix.}
  \label{tab:methods2}
  
  \begin{minipage}[t]{0.48\textwidth}
    \centering
    \caption*{(a) LLaMA3.2-3.2B Results}
    \resizebox{\textwidth}{!}{
    \begin{tabular}{lcccc}
    \hline
    Prune\% & Method & \#Params & PPL $\downarrow$ & Zero-shot A Avg. $\uparrow$ \\
    \hline
    - & - & 3.2B & 10.26 & 54.30 \\
    \hline
    \multirow{3}{*}{20\%}
    & LLMPruner & 2.56B & 22.07 & 52.24 \\
    & LoRAPrune & 2.57B & 19.03 & 50.30 \\
    & \cellcolor{myblue}SDMPrune & \cellcolor{myblue}2.56B & \cellcolor{myblue}\textbf{18.8} & \cellcolor{myblue}\textbf{53.37} \\
    \hline
    \multirow{3}{*}{30\%}
    & LLMPruner & 2.25B & 30.12 & 48.68 \\
    & LoRAPrune & 2.25B & 27.31 & 48.36 \\
    & \cellcolor{myblue}SDMPrune & \cellcolor{myblue}2.25B & \cellcolor{myblue}\textbf{27.24} & \cellcolor{myblue}\textbf{51.03} \\
    \hline
    \end{tabular}
    }
  \end{minipage}
  \hfill
  \begin{minipage}[t]{0.48\textwidth}
    \centering
    \caption*{(b)LLaMA2-7B Results}
    \resizebox{\textwidth}{!}{
    \begin{tabular}{lcccc}
    \hline
    Prune\% & Method & \#Params & PPL $\downarrow$ & Zero-shot A Avg. $\uparrow$ \\
    \hline
    - & - & 6.7B & 7.18 & 58.33 \\
    \hline
    \multirow{3}{*}{20\%}
    & LLMPruner & 5.39B & 11.59 & 56.14 \\
    & LoRAPrune & 5.39B & 11.02 & 56.69 \\
    & \cellcolor{myblue}SDMPrune & \cellcolor{myblue}5.38B & \cellcolor{myblue}\textbf{10.48} & \cellcolor{myblue}\textbf{58.21} \\
    \hline
    \multirow{3}{*}{30\%}
    & LLMPruner & 4.71B & 14.76 & 51.69 \\
    & LoRAPrune & 4.72B & \textbf{12.81} & 54.62 \\
    & \cellcolor{myblue}SDMPrune & \cellcolor{myblue}4.71B & \cellcolor{myblue}12.87 & \cellcolor{myblue}\textbf{55.57} \\
    \hline
    \end{tabular}
    }
  \end{minipage}
  \vspace{-1em}
\end{table}

\begin{table*}[ht]
  \centering
  \caption{Zero-shot performance of other small-scale LLMs and ours. Our pruned model (SDMPrune) is finetuned in a full-parameter manner on the LaMini-instruction dataset for 3 epochs.
  }
  \resizebox{\textwidth}{!}
  {
  \begin{tabular}{l|c|ccccccccccc}
      \hline
      \textbf{Model Name}
      & \textbf{\#Params} & \textbf{ARCc} & \textbf{ARCe}& \textbf{BOOLQ}
      & \textbf{Crows} & \textbf{OBQA}
      & \textbf{PIQA} & \textbf{Race} & \textbf{SiQA} & \textbf{TfQA}& \textbf{Wino}
      & \textbf{Average}
      \\
      \hline
      \hline
      % Llama3.2-1.2B \cite{grattafiori2024llama3herdmodels}&1.2B&37.1&60.7&64.0&62.6&37.6&74.2&37.6&42.9&37.7&60.4&\textbf{51.5} \\
      % \hline
      ShearedLLaMA1.3B \cite{xia2024shearedllamaacceleratinglanguage}&1.3B&29.1&54.4&62.0&63.7&34.4&73.4&36.3&41.3&36.8&58.1&49.0
      \\
      \hline
      TinyLLaMA1.1B \cite{zhang2024tinyllama}&1.1B&30.1&55.3&57.8&62.3&36.0&73.3&36.5&40.6&37.6&59.1&48.9
      \\
      \hline
      Pythia1.0B \cite{biderman2023pythia}&1.1B&26.9&49.0&60.9&60.2&31.4&69.3&32.8&39.8&\textbf{40.5}&53.6&46.4
      \\
      \hline
      Falcon1.3B \cite{almazrouei2023falcon}&1.3B&31.5&57.5&61.5&61.8&35.8&74.6&36.2&41.1&35.8&61.2&49.7
      \\
      \hline
      MobileLLM1.0B \cite{liu2024mobilellm}&1.0B&33.5&58.5&65.6&60.4&\textbf{36.6}&73.6&34.6&41.3&38.3&\textbf{63.3}&50.6
      \\
      \hline
      Openelm1.1B \cite{mehta2024openelm}&1.1B&32.3&55.4&63.6&63.6&36.2&\textbf{75.6}&36.5&42.8&37.0&61.7&50.5
      \\
      \hline
      Opt1.3B \cite{zhang2022opt}&1.3B& 27.8&51.2&57.2&\textbf{65.8}&32.6&70.9&34.2&40.4&38.7&59.4&47.8 \\
      \hline
      MobiLLaMA1.2B \cite{thawakar2024mobillama}&1.2B&31.8&56.5&60.3&64.1&34.8&74.8&34.9&42.0&35.2&59.3&49.4 \\
      \hline
      \cellcolor{myblue} SDMPrune (ours, ratio=20\%) 
      &\cellcolor{myblue}\textbf{1.0B}
      &\cellcolor{myblue}\textbf{35.0}
      &\cellcolor{myblue}\textbf{59.3}
      &\cellcolor{myblue}\textbf{72.7}
      &\cellcolor{myblue}60.5
      &\cellcolor{myblue}34.6
      &\cellcolor{myblue}72.4
      &\cellcolor{myblue}\textbf{37.0}
      &\cellcolor{myblue}\textbf{44.2}
      &\cellcolor{myblue}39.7
      &\cellcolor{myblue}58.5
      &\cellcolor{myblue}\textbf{51.4} \\
      \hline
    \end{tabular}
  }
  % \vspace{-0.5cm}
  \label{tab:models}
\end{table*}
\subsection{Comparison of Existing Methods}
\label{sec:comparewithmethods}

To facilitate a more rigorous and comparative analysis with existing pruning methods, we applied structural pruning to the LLaMA3.2 and LLaMA2 models. 
This process yields many models variants with different model size, facilitating an evaluation of the trade-offs between model compression and performance degradation.
Each model was finetuned using LoRA on the LaMini-instruction dataset for two epochs.

It is noteworthy that the vocabulary size of the LLaMA3.2 series models is four times larger than that of previous models. 
Even though the token embedding layer shares weights with the classification head, this component still accounts for up to 21\% of the total model parameters. 
While most prior work primarily focuses on proportional pruning of parameters within the transformer blocks, our implementation adopts the final model size as the pruning objective.

We have conducted a comparison of SDMPrune against several existing methods, which include Magnitude \cite{han2015learningweightsconnectionsefficient}, wanda \cite{sun2024simpleeffectivepruningapproach}, LLMPruner \cite{ma2023llm}, Compresso \cite{guo2023compressostructuredpruningcollaborative}, LoRAPrune \cite{zhang2024loraprunestructuredpruningmeets}.
Among these methods, Magnitude and wanda primarily concentrate on pruning by targeting activation values and weights, respectively. 
Compresso, on the other hand, accomplishes pruning through the learning of a set of binary masks. 
Meanwhile, LLMPruner and LoRAPrune employ gradient information to facilitate the pruning process.

Table ~\ref{tab:methods} and Table ~\ref{tab:methods2} present a comprehensive comparative evaluation of our pruning methodology in contrast to other structural pruning methods.
The experimental results of our SDMPrune under structural pruning conditions exhibit superior performance compared to alternative methods. 
Notably, pruning methods incorporating gradient information significantly outperform those based solely on weight and activation values, which indicates that task-specific feedback more accurately captures the importance distribution of model parameters.
The experimental results demonstrate that our SDMPrune method outperforms both LoRAPrune and LLMPruner. 
This suggests that gradient information, which focuses on preserving the full prediction probability distribution, is more accurate than simply enhancing label token prediction accuracy. 
SDMPrune demonstrates nearly linear acceleration proportional to the parameter reduction ratio, which validates its effectiveness in LLM compression tasks. 
Detailed acceleration results are presented in the Appendix.
\subsection{Comparison of Existing Models}
\label{sec:comparewithmodels}

We conducted SDMPrune methods with pruning ratio of 20\% applied to the parameters of each layer, followed by full-parameter finetuning on the LaMini-instruction dataset for three epochs.

As shown in Table~\ref{tab:models}, among models of the same size, the model pruned using SDMPrune achieves very competitive performance after just three epochs of finetuning, with only a marginal performance degradation of 0.1\% compared to the original model.
This negligible performance gap underscores the efficacy of SDMPrune in achieving model compression, effectively reducing computational overhead, while maintaining the model's performance integrity.
% This observation further suggests that the hidden layer neurons in the MLP of large models exhibit considerable redundancy. 
% Consequently, a highly effective strategy for achieving a high-performance model is to apply structured pruning to a well-trained, pre-existing larger model.
\begin{table*}[t]
  \centering
  \caption{Zero-shot performance difference before and after using the  MLP-only pruning strategy in our method.}
  \resizebox{\textwidth}{!}
  {
    \begin{tabular}{l|l|lllllllllll}
      \hline
      \textbf{Ratio}
      & \textbf{Method} 
      & \textbf{ARCc} & \textbf{ARCe} & \textbf{BOOLQ} &\textbf{Crows} & \textbf{OBQA}
      & \textbf{PIQA} & \textbf{Race} & \textbf{SiQA} & \textbf{TfQA}  & \textbf{Wino}  
      & \textbf{Average$\uparrow$}
      \\
      \hline\hline
      \multirow{4}{*}{20\%} & attn\&mlp w/o tune 
      &26.74 &48.73 &58.35 &54.39 &33.00 &68.33 &31.25 &40.27&43.35&55.30&45.97 \\
       & attn\&mlp w/ tune 
       &30.63 &53.97 &64.24 &55.24 &32.30 &69.32 &32.95 &40.60 &40.53 &56.07 &47.59 \\
       & mlp w/o tune 
       &27.90\textcolor{blue}{(+1.16)} &48.91\textcolor{blue}{(+0.18)} &60.12\textcolor{blue}{(+1.77)} 
       &58.20\textcolor{blue}{(+3.81)} &32.80\textcolor{red}{(-0.20)} &68.11\textcolor{red}{(-0.22)} 
       &33.21\textcolor{blue}{(+1.96)} &40.69\textcolor{blue}{(+0.42)} &44.49\textcolor{blue}{(+1.14)} 
       &55.64\textcolor{blue}{(+0.34)} &47.01\textcolor{blue}{(+1.04)} \\
       & mlp w/ tune 
       &31.14\textcolor{blue}{(+0.51)} &55.22\textcolor{blue}{(+1.25)} &67.58\textcolor{blue}{(+3.34)} 
       &57.84\textcolor{blue}{(+2.60)} &32.40\textcolor{blue}{(+0.10)} &70.29\textcolor{blue}{(+0.97)} 
       &35.41\textcolor{blue}{(+2.46)} &42.73\textcolor{blue}{(+2.13)} &40.20\textcolor{red}{(-0.33)} 
       &56.59\textcolor{blue}{(+0.52)} &48.94\textcolor{blue}{(+1.35)} \\
      \hline\hline
      \multirow{4}{*}{30\%} & attn\&mlp w/o tune &23.96 &41.03 
      &55.35 &50.49 &27.00 &61.50 &30.27 &37.46 &44.97 &52.65 &42.47  \\
       & attn\&mlp w/ tune 
       &24.50 &44.30&62.33 &52.96 &27.70 &63.97 &32.87 &38.61 &42.09 &52.72 &44.20 \\
       & mlp w/o tune 
       &24.40\textcolor{blue}{(+0.44)} &41.25\textcolor{blue}{(+0.22)} &59.05\textcolor{blue}{(+3.7)} 
       &52.95\textcolor{blue}{(+2.46)} &27.00\textcolor{blue}{(+0)} &63.28\textcolor{blue}{(+1.78)} 
       &30.81\textcolor{blue}{(+0.54)} &38.43\textcolor{blue}{(+0.97)} &47.31\textcolor{blue}{(+2.34)} 
       &55.64\textcolor{blue}{(+2.99)} &44.01\textcolor{blue}{(+1.54)} \\
       & mlp w/ tune 
       &28.50\textcolor{blue}{(+4.00)} &47.47\textcolor{blue}{(+3.17)} &64.68\textcolor{blue}{(+2.35)} 
       &56.89\textcolor{blue}{(+3.93)} &29.00\textcolor{blue}{(+1.30)} &66.32\textcolor{blue}{(+2.35)} 
       &33.21\textcolor{blue}{(+0.34)} &40.84\textcolor{blue}{(+2.23)} &42.35\textcolor{blue}{(+0.26)} 
       &54.70\textcolor{blue}{(+1.98)} &46.40\textcolor{blue}{(+2.20)} \\
       \hline\hline
      \multirow{4}{*}{40\%} 
        & attn\&mlp w/o tune 
        &23.01 &32.64 
        &56.32 &49.37 &24.00 &58.41 &27.40 &35.72 &45.13 &49.26 &40.13  \\
        & attn\&mlp w/ tune 
        &25.41 &41.90&63.92 &51.82 &24.70 &62.44 &31.93 &37.13 &44.12 &50.95 &43.43 \\
        & mlp w/o tune 
        &23.04\textcolor{blue}{(+0.03)} &33.80\textcolor{blue}{(+1.16)} &61.01\textcolor{blue}{(+4.69)} 
        &51.04\textcolor{blue}{(+1.67)} &25.20\textcolor{blue}{(+1.20)} &58.32\textcolor{red}{(-0.09)} 
        &28.23\textcolor{blue}{(+0.83)} &37.15\textcolor{blue}{(+1.43)} &46.30\textcolor{blue}{(+1.17)} 
        &49.72\textcolor{blue}{(+0.46)} &41.38\textcolor{blue}{(+1.25)} \\
        & mlp w/ tune 
        &26.02\textcolor{blue}{(+0.61)} &42.63\textcolor{blue}{(+0.73)} &65.38\textcolor{blue}{(+1.46)} 
        &52.59\textcolor{blue}{(+0.77)} &25.60\textcolor{blue}{(+0.40)} &63.44\textcolor{blue}{(+1.00)} 
        &32.25\textcolor{blue}{(+0.32)} &38.74\textcolor{blue}{(+1.61)} &43.30\textcolor{red}{(-0.82)} 
        &52.17\textcolor{blue}{(+1.22)} &44.21\textcolor{blue}{(+0.78)} \\
        \hline\hline
      \end{tabular}
  }
  \label{tab:differentPrune}
\end{table*}
\begin{table*}[t]
  \centering
  \begin{minipage}[t]{0.6\textwidth} % 左侧表格
    \centering
    \caption{Zero-shot performance difference before and after using the MLP-only pruning strategy in existing methods.}
    \resizebox{\textwidth}{!}
    {
      \begin{tabular}{l|l|c|c|l}
        \hline
        \textbf{Method} & \textbf{Ratio} & \textbf{MLP\&ATTN Zero-shot Avg.} & \textbf{MLP Zero-shot Avg.} & \textbf{Increase} \\ \hline
        \multirow{3}{*}{wanda} & 20\% & 46.27 & 47.56 & \textbf{+1.29} \\ \cline{2-5}
                               & 30\% & 44.01 & 45.09 & \textbf{+1.08} \\ \cline{2-5}
                               & 40\% & 42.08 & 43.14 & \textbf{+1.06} \\ \hline
        \multirow{3}{*}{magnitude} & 20\% & 45.12 & 46.79 & \textbf{+1.67} \\ \cline{2-5}
                                   & 30\% & 43.27 & 44.61 & \textbf{+1.34} \\ \cline{2-5}
                                   & 40\% & 41.95 & 43.14 & \textbf{+1.19} \\ \hline
        \multirow{3}{*}{SDMPrune}  & 20\% & 47.59 & 48.94 & \textbf{+1.35} \\ \cline{2-5}
                                   & 30\% & 44.20 & 46.40 & \textbf{+2.20} \\ \cline{2-5}
                                   & 40\% & 43.43 & 44.21 & \textbf{+0.78} \\ \hline
      \end{tabular}
    }
    \label{tab:method_mlp}
  \end{minipage}
  \hfill % 填充空白，使两个 minipage 分开
  \begin{minipage}[t]{0.35\textwidth} % 右侧表格
    \centering
    \caption{Zero-shot performance differences of the model finetuned on different parts after pruning using our method.}
    \resizebox{\textwidth}{!}
    {
      \begin{tabular}{c|c|c|c}
        \hline
        \textbf{Ratio} & \textbf{MLP} & \textbf{ATTN} &\textbf{Zero-shot Avg.$\uparrow$} \\ \hline
        \multirow{2}{*}{20\%} & $\checkmark$ & $\checkmark$ & 48.94 \\
                              & $\checkmark$ &$\times$& 48.92 \\ \hline
        \multirow{2}{*}{30\%} & $\checkmark$ &$\checkmark$&46.40 \\
                              & $\checkmark$ &$\times$& 46.10 \\ \hline
        \multirow{2}{*}{40\%} & $\checkmark$ &$\checkmark$&44.21 \\
                              & $\checkmark$ &$\times$& 43.65 \\ \hline
      \end{tabular}
    }
    \label{tab:differentFinetune}
  \end{minipage}
  \vspace{-0.5cm}
\end{table*}
\subsection{Ablation Study}
\label{sec:ablation}
\subsubsection{Impact of MLP-only Pruning}
The motivation for pruning only MLP layers stems from their distinct role in transformer-based architectures.
While attention layers capture long-range dependencies and token interactions, MLP layers primarily perform feature transformation and refinement.
Previous approaches that pruned both components risked degrading sequence-level feature extraction and reducing performance recoverability.
By strategically pruning MLP layers while maintaining the attention mechanism's structured integrity, we ensure robust sequence modeling capabilities, thereby enhancing model recoverability.

To analyze the impact of MLP-only pruning on the performance recovery of pruned models, we conducted three sets of experiments following the experimental setup described in Section ~\ref{sec:comparewithmethods}:

\textbf{Prune Different Parts.}
We performed two sets of pruning experiments—one pruning both the attention layers and MLP layers simultaneously, and the other exclusively pruning the MLP layers. 
Both sets of pruned models were then LoRA finetuned.
Table~\ref{tab:differentPrune} presents the impact of pruning different components on experimental results. 
Regardless of whether the model is evaluated without finetuning or with finetuning, pruning only MLP parameters while preserving attention layers yields superior performance. 
This suggests that selective MLP pruning better preserves linguistic capabilities, enabling better recovery with minimal fine-tuning. 
In contrast, pruning attention layers disrupts token interaction modeling, impairing convergence and overall performance.

\textbf{Application of MLP-only Pruning to Previous Methods.}
We implemented the MLP-only pruning strategy in existing methods and evaluated the performance of the finetuned models both before and after applying this strategy, emphasizing the effectiveness of selectively pruning MLP parameters.
Table~\ref{tab:method_mlp} reports the impact of applying the MLP pruning strategy on the Wanda and Magnitude methods. 
Finetuned models at various pruning ratios demonstrate notable performance improvements, achieving results comparable to those obtained with existing gradient-based pruning methods.

\textbf{Finetune Different Parts.}
For models pruned by SDMPrune, we compare performance across different finetuned modules. 
Specifically, we compare finetuning only MLP layers against finetuning both MLP and attention layers.
As shown in Table~\ref{tab:differentFinetune}, finetuning only MLP layers achieves performance nearly identical to that of both MLP and attention layers finetuning at a 20\% pruning ratio. 
At higher pruning ratios, the performance difference increases to 0.4\%, which is still acceptable.
This suggests that keeping the attention layer parameters unchanged preserves the model's ability to integrate and extract sequence features, while finetuning the MLP layers enables the adaptation of these features by projecting them into the appropriate feature space under the influence of the finetuning dataset. 
This experiment verifies our hypothesis and indirectly proves the effectiveness of SDMPrune for LLMs.

\subsubsection{Impact Of Distillation Loss}
We performed a series of experiments by systematically adjusting the temperature parameters and loss ratios to comprehensively analyze their impact on the results.

\begin{figure*}[h]
  \includegraphics[width=\textwidth]{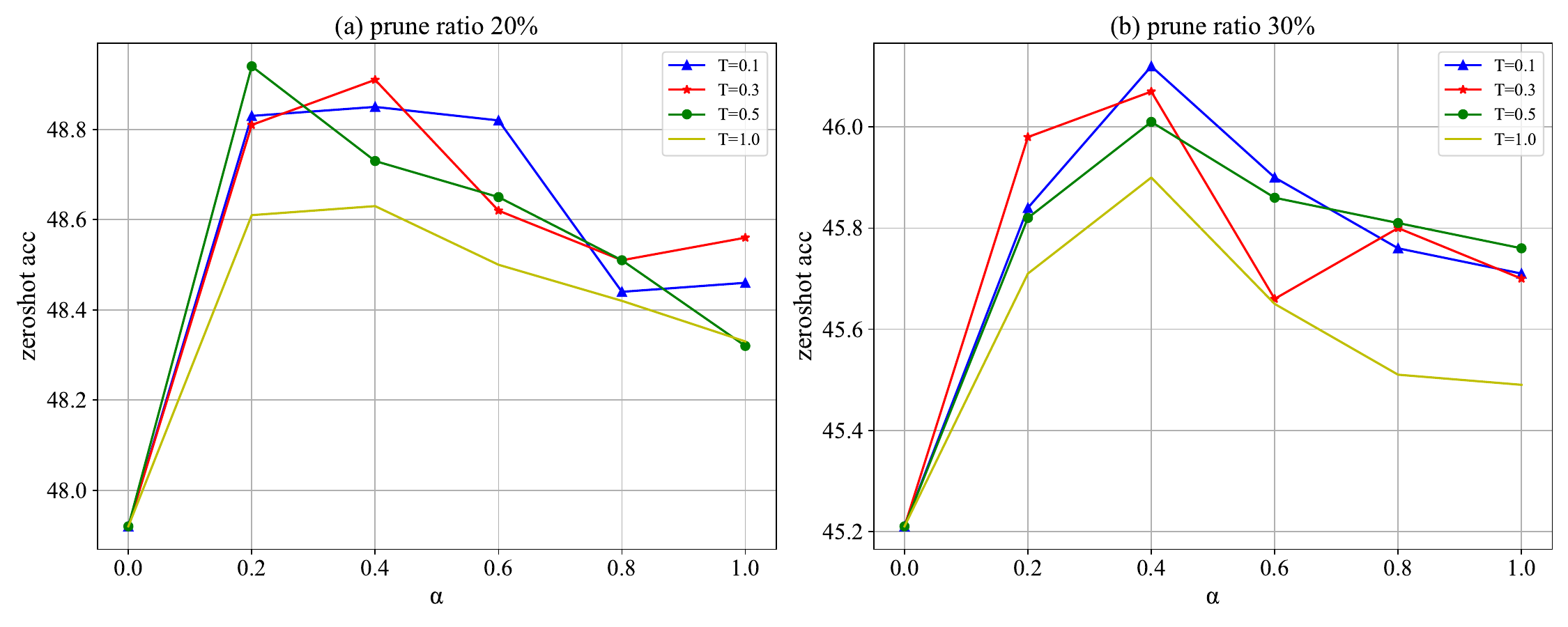}
  \caption{The model's performance under different $t$ and $\alpha$ configurations of the SDMPrune strategy.}
  \label{fig:hyperparam}
\end{figure*}

When $\alpha$ is set to 0, no self-distillation loss is introduced.
As observed in Figure ~\ref{fig:hyperparam}, the incorporation of self-distillation loss consistently demonstrates superior performance over the baseline model pruned exclusively with the one-hot label encoding loss.
This demonstrates that the introduction of self-distillation loss enables the model to comprehensively preserve its language prediction capabilities during pruning.
The more fine-grained gradient information derived from this loss function more accurately reflects the importance of neurons, thereby ensuring the effectiveness of the pruning method.

In our experiments, when $t$ is below 0.5, the performance of the models with different $\alpha$ shows little difference, whereas at a temperature of 1, the performance significantly deteriorates. 
This indicates that focusing on potential candidate tokens during pruning indeed improves pruning effectiveness.
However, excessive attention to non-candidate tokens can also prevent the model from achieving optimal performance.

\vspace{-1em}

\section{Conclusion}
\label{sec:conclusion}
In this paper, we propose a novel method for pruning LLMs that outperforms existing pruning methods.
Specifically, we introduce a self-distillation loss to enable the pruned model to retain the complete prediction probability distribution of the original model.
Building on this loss function, we conduct experimental analyses and propose pruning only the MLP layers while retaining the complete attention layers, ensuring the pruned model maintains its sequence processing capability.
Extensive evaluations demonstrate the superiority of SDMPrune over existing pruning methods, while ablation studies validate the effectiveness of our two key innovations.

\textbf{Limitations}.
However, SDMPrune introduces teacher model-assisted pruning, which results in increased memory usage and computational overhead.
Looking ahead, we aim to explore lightweight implementation of distillation and further enhance SDMPrune's pruning performance at higher compression rates.

% \newpage
{
    \small
    \bibliographystyle{plain}
    % \bibliography{refer}
    \bibliography{paper.bbl}
}

\newpage
\appendix
% \section{Cholesky Decomposition}

\section{Implementation Details.}
Table ~\ref{tab:implementation_details} presents the detailed experimental settings for structural pruned model tuning. 
Both Section 3.2 and the ablation study were conducted on LLaMA3.2-1.2B, following identical pruning settings to those specified for Llama3.2-1.2B in Table ~\ref{tab:implementation_details}.
During the finetuning process, we preprocess the data into the packing inputs without Cross-Contamination Attention \cite{Gultekin_Functionary}.
With this optimization approach, our experimental setup involves a batch size of 1 per GPU, a sample length of 4096,  gradient accumulation over 12 steps, which is nearly equivalent to a batch size of 240.

\begin{table}[ht]
    \centering
    \caption{Implementation Details of the structural pruned LLMs.}
    \resizebox{\textwidth}{!}{
        \begin{tabular}{c|c|c|c|c|c|c|c|c|c}
        \hline
        \textbf{Model} & \textbf{Ratio} & \textbf{Pack} & \textbf{lr} & \textbf{beta1} & \textbf{beta2} & \textbf{Weight decay} & \textbf{Precision} & \textbf{Batch size} & \textbf{Device} \\
        \hline
        \multirow{3}{*}{LLaMA3.2-1.2B}
        & 20\% & yes & $1\times10^{-4}$ & 0.9 & 0.999 & 0 & BF16 & 240 & 4$\times$4090 \\
        & 30\% & yes & $1\times10^{-4}$ & 0.9 & 0.999 & 0 & BF16 & 240 & 4$\times$4090 \\
        & 40\% & yes & $1\times10^{-4}$ & 0.9 & 0.999 & 0 & BF16 & 240 & 4$\times$4090 \\
        \hline
        \multirow{2}{*}{LLaMA3.2-3.2B}
        & 20\% & yes & $5\times10^{-4}$ & 0.9 & 0.999 & 0 & BF16 & 240 & 4$\times$4090 \\
        & 30\% & yes & $5\times10^{-4}$ & 0.9 & 0.999 & 0 & BF16 & 240 & 4$\times$4090 \\
        \hline
        \multirow{2}{*}{LLaMA2-7B}
        & 20\% & no & $1\times10^{-4}$ & 0.9 & 0.999 & 0 & BF16 & 256 & 8$\times$A6000 \\
        & 30\% & no & $1\times10^{-4}$ & 0.9 & 0.999 & 0 & BF16 & 256 & 8$\times$A6000 \\
        \hline
        \end{tabular}
    }
    \label{tab:implementation_details}
\end{table}

\section{More Detailed Evaluation Results}
\textbf{Detailed evaluation results of pruned LLaMA3.2-3.2B and LLaMA2-7B models.}
Tables ~\ref{tab:methods_3B} and Tables ~\ref{tab:methods_7B} present detailed model evaluation results, 
serving as an expanded version of the data in Table ~\ref{tab:methods2}.

\begin{table*}[ht]
    \centering
    \caption{Zero-shot performance and Perplexity of the compressed LLaMA3.2-3.2B model.
    }
    \resizebox{\textwidth}{!}
    {
    \begin{tabular}{l|l|c|ccccccccccc}
        \hline
        \textbf{Ratio}
        & \textbf{Method} & \textbf{PPL$\downarrow$}
        & \textbf{BOOLQ} & \textbf{Crows} &\textbf{OBQA}&  \textbf{PIQA} 
        & \textbf{RACE} & \textbf{SiQA}  & \textbf{TfQA} 
        & \textbf{Average$\uparrow$}
        \\
        \hline\hline
        0\% & LLaMA3.2-3.2B&10.26&73.67&63.33&40.60&77.86&	38.85&46.67&39.15&54.30\\
        \hline\hline
        \multirow{6}{*}{20\%} 
        &Magnitude\cite{han2015learningweightsconnectionsefficient} &27.43& 68.06 & 55.82 & 31.80 & 70.92 & 34.27 & 41.62 & 42.76 & 49.32 \\
        &Wanda\cite{sun2024simpleeffectivepruningapproach} & 24.21 &69.12 & 56.38 & 33.20 & 71.22 & 35.03 & 41.27 & 43.24 & 49.92 \\
        &LLMPruner\cite{ma2023llm} & 22.07 & 74.65 & 58.86 & 33.80 & 72.09 & \textbf{40.10} & 43.65 & 42.56 & 52.24 \\
        &LoRAPrune\cite{zhang2024loraprunestructuredpruningmeets} & 19.03 & 68.81 & 57.94 & 33.10 & 71.42 & 35.31 & 42.94 & 42.61 & 50.30 \\
        & \cellcolor{myblue}SDMPrune(ours) & \cellcolor{myblue}\textbf{18.8} & \cellcolor{myblue}\textbf{75.29}
        & \cellcolor{myblue}\textbf{59.33} & \cellcolor{myblue}\textbf{36.20} 
        & \cellcolor{myblue}\textbf{73.61} & \cellcolor{myblue}38.85 & \cellcolor{myblue}\textbf{44.58} &\cellcolor{myblue}\textbf{45.75} & \cellcolor{myblue}\textbf{53.37} \\
        \hline\hline
        \multirow{6}{*}{30\%} 
        & Magnitude \cite{han2015learningweightsconnectionsefficient}
        &36.01&64.65 & 54.76 & 30.10 & 63.02 & 33.24 & 40.77 & 40.83 & 46.77  \\
        & Wanda \cite{sun2024simpleeffectivepruningapproach}
        &34.45&64.32 & 54.89 & 29.40 & 66.15 & 34.02 & 40.20 & 41.20 & 47.17 \\
        & LLMPruner \cite{ma2023llm}
        & 30.12 & 66.21 & 54.92 & 32.80 & 66.21 & \textbf{37.13} & 41.25 & 42.27 & 48.68 \\
        & LoRAPrune \cite{zhang2024loraprunestructuredpruningmeets}
        & 27.31 &65.44 & 53.87 & \textbf{33.80} & 69.53 & 33.97 & 41.20 & 40.74 & 48.36 \\
        & \cellcolor{myblue}SDMPrune(ours)
        &\cellcolor{myblue}\textbf{27.24}&\cellcolor{myblue}\textbf{72.78} & \cellcolor{myblue}\textbf{58.14}
        & \cellcolor{myblue}30.40 & \cellcolor{myblue}\textbf{70.67} & \cellcolor{myblue}36.84
        & \cellcolor{myblue}\textbf{43.30} & \cellcolor{myblue}\textbf{45.10} & \cellcolor{myblue}\textbf{51.03} \\
        \hline\hline
      \end{tabular}
    }
    \label{tab:methods_3B}
\end{table*}

\begin{table*}[ht]
    \centering
    \caption{Zero-shot performance and Perplexity of the compressed LLaMA2-7B model.}
    \resizebox{\textwidth}{!}
    {
    \begin{tabular}{l|l|c|ccccccccccc}
        \hline
        \textbf{Ratio}
        & \textbf{Method} & \textbf{PPL$\downarrow$}
        & \textbf{ARCc} & \textbf{ARCe} & \textbf{BOOLQ} &\textbf{Crows} & \textbf{OBQA}
        & \textbf{PIQA} & \textbf{Race} & \textbf{SiQA} & \textbf{TfQA}  & \textbf{Wino}  
        & \textbf{Average$\uparrow$}
        \\
        \hline\hline
        0\% & LLaMA2-7B&7.18&45.14 & 73.82 & 79.36 & 67.44 & 44.20 & 78.73 & 40.10 & 46.52 &38.75& 69.30 & 58.33 \\
        \hline\hline
        \multirow{6}{*}{20\%} 
        & Magnitude \cite{han2015learningweightsconnectionsefficient} 
        &16.01& 37.82 & 66.29 & 75.62 & 57.82 & 37.80 & 72.51 & 37.82 & 46.02 & 44.36 & 63.82 & 53.99 \\
        & Wanda \cite{sun2024simpleeffectivepruningapproach}
        &14.27 &39.82 & 68.72 & 76.21 & 59.83 & 37.70 & 74.12 & 39.65 & 44.82 & 44.01 & 63.61 & 54.85 \\
        & LLMPruner \cite{ma2023llm}
        &11.59 &40.36 & 70.12 & 80.21 & 61.66 & 38.80 & 75.84 & 39.04 & 47.13 & 43.91 & 64.33 & 56.14\\
        & LoRAPrune \cite{zhang2024loraprunestructuredpruningmeets}
        &12.81 &41.64 & 71.00 & \textbf{81.71} & 58.67 & 41.40 & 76.71 & 40.38 & 44.03 & 65.88 & 65.88 & 56.69 \\
        & \cellcolor{myblue}SDMPrune(ours)
        &\cellcolor{myblue}\textbf{10.48} &\cellcolor{myblue}\textbf{43.94}&\cellcolor{myblue}\textbf{72.31}
        &\cellcolor{myblue}81.65&\cellcolor{myblue}\textbf{62.08}&\cellcolor{myblue}\textbf{42.00}
        &\cellcolor{myblue}\textbf{77.04}&\cellcolor{myblue}\textbf{41.34}&\cellcolor{myblue}\textbf{48.52}
        &\cellcolor{myblue}\textbf{44.92}&\cellcolor{myblue}\textbf{68.35}&\cellcolor{myblue}\textbf{58.21}
        \\
        \hline\hline
        \multirow{6}{*}{30\%} 
        & Magnitude \cite{han2015learningweightsconnectionsefficient}
        &19.02 &35.61 & 61.73 & 72.01 & 59.83 & 36.80 & 70.32 & 35.77 & 45.87 & 42.16 & 55.98 & 51.61  \\
        & Wanda \cite{sun2024simpleeffectivepruningapproach}
        &18.37 &36.47 & 62.07 & 73.20 & 60.62 & 37.10 & 71.01 & 35.21 & 44.12 & 43.65 & 57.01 & 52.05 \\
        & LLMPruner \cite{ma2023llm}
        &14.76 &38.01 & 64.80 & 75.63 & \textbf{62.25} & 36.40 & 73.39 & 35.69 & 47.32 & 42.25 & 62.88 & 53.86 \\
        & LoRAPrune \cite{zhang2024loraprunestructuredpruningmeets}
        &\textbf{12.81} &38.64 & 65.14 &74.06 & 61.42  & \textbf{37.40} & 72.92 & 39.04 & 46.26 & \textbf{44.77} & \textbf{66.54} & 54.62  \\
        & \cellcolor{myblue}SDMPrune(ours)
        &\cellcolor{myblue}12.87 &\cellcolor{myblue}\textbf{39.59} & \cellcolor{myblue}\textbf{67.93}
        & \cellcolor{myblue}\textbf{80.37} & \cellcolor{myblue}58.50 & \cellcolor{myblue}37.20 
        & \cellcolor{myblue}\textbf{75.24} & \cellcolor{myblue}\textbf{40.00} & \cellcolor{myblue}\textbf{47.80}
        & \cellcolor{myblue}43.70 & \cellcolor{myblue}65.35 & \cellcolor{myblue}\textbf{55.57} \\
        \hline\hline
      \end{tabular}
    }
    \label{tab:methods_7B}
\end{table*}

\newpage

\textbf{Acceleration evaluation results of pruned models.}
Table ~\ref{tab:acceleration} records the acceleration results of the pruned models. 
Since the actual runtime can be affected by factors such as device bandwidth and computational efficiency, 
we use the model's computational workload (Macs) to represent its acceleration performance which runs on sequence with a length of 256.

\begin{table}[h]
    \centering
    \caption{Acceleration results of the structural pruned LLMs.}
    \resizebox{0.7\textwidth}{!}{
        \begin{tabular}{c|c|c|c|c|c}
        \hline
        \textbf{Model} & \textbf{\#Params} & \textbf{Ratio} & \textbf{Zero-shot avg$\uparrow$} & \textbf{Mac} & \textbf{Acceleration} \\
        \hline
        \multirow{3}{*}{LLaMA3.2-1.2B}
        & 1.2B & 0 & 51.47 & 158.18G & -- \\
        & 1.0B & 20\% & 48.94 & 124.66G & 21.2\% \\
        & 0.8B & 30\% & 46.40 & 108.35G & 31.5\% \\
        \hline
        \multirow{3}{*}{LLaMA3.2-3.2B}
        & 3.2B & 0 & 54.30 & 411.21G & -- \\
        & 2.6B & 20\% & 53.37 & 330.35G & 19.7\% \\
        & 2.3B & 30\% & 51.07 & 289.26G & 29.7\% \\
        \hline
        \multirow{3}{*}{LLaMA2-7B}
        & 6.7B & 0 & 58.33 & 845.71G & -- \\
        & 5.4B & 20\% & 58.21 & 665.57G & 21.3\% \\
        & 4.7B & 30\% & 55.57 & 588.61G & 30.4\% \\
        \hline
        \end{tabular}
    }
    \label{tab:acceleration}
\end{table}

\end{document}